\documentclass[sigconf]{acmart}
\AtBeginDocument{%
  \providecommand\BibTeX{{%
    \normalfont B\kern-0.5em{\scshape i\kern-0.25em b}\kern-0.8em\TeX}}}

\setcopyright{acmcopyright}

\copyrightyear{2021} 
\acmYear{2021} 
\setcopyright{acmcopyright}\acmConference[SIGIR '21]{Proceedings of the 44th International ACM SIGIR Conference on Research and Development in Information Retrieval}{July 11--15, 2021}{Virtual Event, Canada}
\acmBooktitle{Proceedings of the 44th International ACM SIGIR Conference on Research and Development in Information Retrieval (SIGIR '21), July 11--15, 2021, Virtual Event, Canada}
\acmPrice{15.00}
\acmDOI{10.1145/3404835.3462925}
\acmISBN{978-1-4503-8037-9/21/07}

\usepackage{multirow}
\usepackage{array}
\makeatletter  
\newif\if@restonecol  
\makeatother

\usepackage[linesnumbered,ruled,vlined]{algorithm2e}
\usepackage{algpseudocode}  
\usepackage{amsmath}  
\usepackage{balance}

\begin{document}
\fancyhead{}
\title{Relational Learning with Gated and Attentive Neighbor Aggregator for Few-Shot Knowledge Graph Completion}

\author{Guanglin Niu}
\authornote{Corresponding authors.}
\affiliation{%
  \institution{DAMO Academy, Alibaba Group}
  \city{Beijing}
  \country{China}
}
\email{guanglin.ngl@alibaba-inc.com}

\author{Yang Li}
\authornotemark[1]
\affiliation{%
  \institution{DAMO Academy, Alibaba Group}
  \city{Beijing}
  \country{China}
}
\email{ly200170@alibaba-inc.com}

\author{Chengguang Tang}
\authornotemark[1]
\affiliation{%
  \institution{DAMO Academy, Alibaba Group}
  \city{Beijing}
  \country{China}
}
\email{chengguang.tcg@alibaba-inc.com}

\author{Ruiying Geng}
\affiliation{%
  \institution{DAMO Academy, Alibaba Group}
  \city{Beijing}
  \country{China}
}
\email{ruiying.gry@alibaba-inc.com}

\author{Jian Dai}
\affiliation{%
  \institution{Alibaba Group}
  \city{Beijing}
  \country{China}
}
\email{yiding.dj@alibaba-inc.com}

\author{Qiao Liu}
\affiliation{%
  \institution{}
  \city{Chengdu}
  \country{China}
}
\email{cnliuqiao@gmail.com}

\author{Hao Wang}
\affiliation{%
  \institution{Alibaba Group}
  \city{Beijing}
  \country{China}
}
\email{qiao.wh@alibaba-inc.com}

\author{Jian Sun}
\affiliation{%
  \institution{DAMO Academy, Alibaba Group}
  \city{Beijing}
  \country{China}
}
\email{jian.sun@alibaba-inc.com}

\author{Fei Huang}
\affiliation{%
  \institution{DAMO Academy, Alibaba Group}
  \city{Seattle}
  \state{WA}
  \country{USA}
}
\email{f.huang@alibaba-inc.com}

\author{Luo Si}
\affiliation{%
  \institution{DAMO Academy, Alibaba Group}
  \city{Seattle}
  \state{WA}
  \country{USA}
}
\email{luo.si@alibaba-inc.com}

\begin{abstract}
  Aiming at expanding few-shot relations' coverage in knowledge graphs (KGs), few-shot knowledge graph completion (FKGC) has recently gained more research interests. Some existing models employ a few-shot relation's multi-hop neighbor information to enhance its semantic representation. However, noise neighbor information might be amplified when the neighborhood is excessively sparse and no neighbor is available to represent the few-shot relation. Moreover, modeling and inferring complex relations of one-to-many (1-N), many-to-one (N-1), and many-to-many (N-N) by previous knowledge graph completion approaches requires high model complexity and a large amount of training instances. Thus, inferring complex relations in the few-shot scenario is difficult for FKGC models due to limited training instances. In this paper, we propose a few-shot relational learning with global-local framework to address the above issues. At the global stage, a novel gated and attentive neighbor aggregator is built for accurately integrating the semantics of a few-shot relation's neighborhood, which helps filtering the noise neighbors even if a KG contains extremely sparse neighborhoods. For the local stage, a meta-learning based TransH (MTransH) method is designed to model complex relations and train our model in a few-shot learning fashion. Extensive experiments show that our model outperforms the state-of-the-art FKGC approaches on the frequently-used benchmark datasets NELL-One and Wiki-One. Compared with the strong baseline model MetaR, our model achieves 5-shot FKGC performance improvements of 8.0\% on NELL-One and 2.8\% on Wiki-One by the metric Hits@10.

\end{abstract}



\maketitle

\section{Introduction}

Knowledge graph (KG) stores rich multi-relational data in a directed graph structure. Many KGs in the real world, such as Freebase~\cite{BGF:Freebase}, YAGO~\cite{suchanek2007yago}, WordNet~\cite{Miller:WordNet}, Wikidata \cite{wikidata} and NELL~\cite{Mitchell:nell}, consist of triple facts, which are represented in the form of (\emph{head entity}, \emph{relation}, \emph{tail entity}). KGs have been introduced to a variety of applications such as information extraction\cite{hoffmann2011knowledge,daiber2013improving}, semantic search~\cite{berant2013semantic,www2017semantic}, dialogue system~\cite{he2017learning, Zhou:Commonsense}, and question answering\cite{zhang2016question,diefenbach2018wdaqua}, to name a few. However, as most KGs are incomplete, it is necessary to conduct knowledge graph completion (KGC) by inferring new facts to improve the use of KGs.

Over the past decade, knowledge graph embedding (KGE) has proven to be powerful for KGC tasks~\cite{KGE:Survey,Analogical, Chami:Hyperbolic}. KGE aims to embed entities and relations into latent, low-dimensional numerical representations. Many KGE techniques have been proposed and are available to KGC tasks, including TransE~\cite{Bordes:TransE}, TransH~\cite{Wang:TransH}, TransR~\cite{Lin-a:TransR}, ComplEx~\cite{Trouillon:ComplEx} and ConvE~\cite{Dettmers:CNN}. In particular, all these approaches assume that 
KGs contain sufficient instances for both entities and relations.

\begin{figure}
  \centering
  \includegraphics[width=0.47\textwidth]{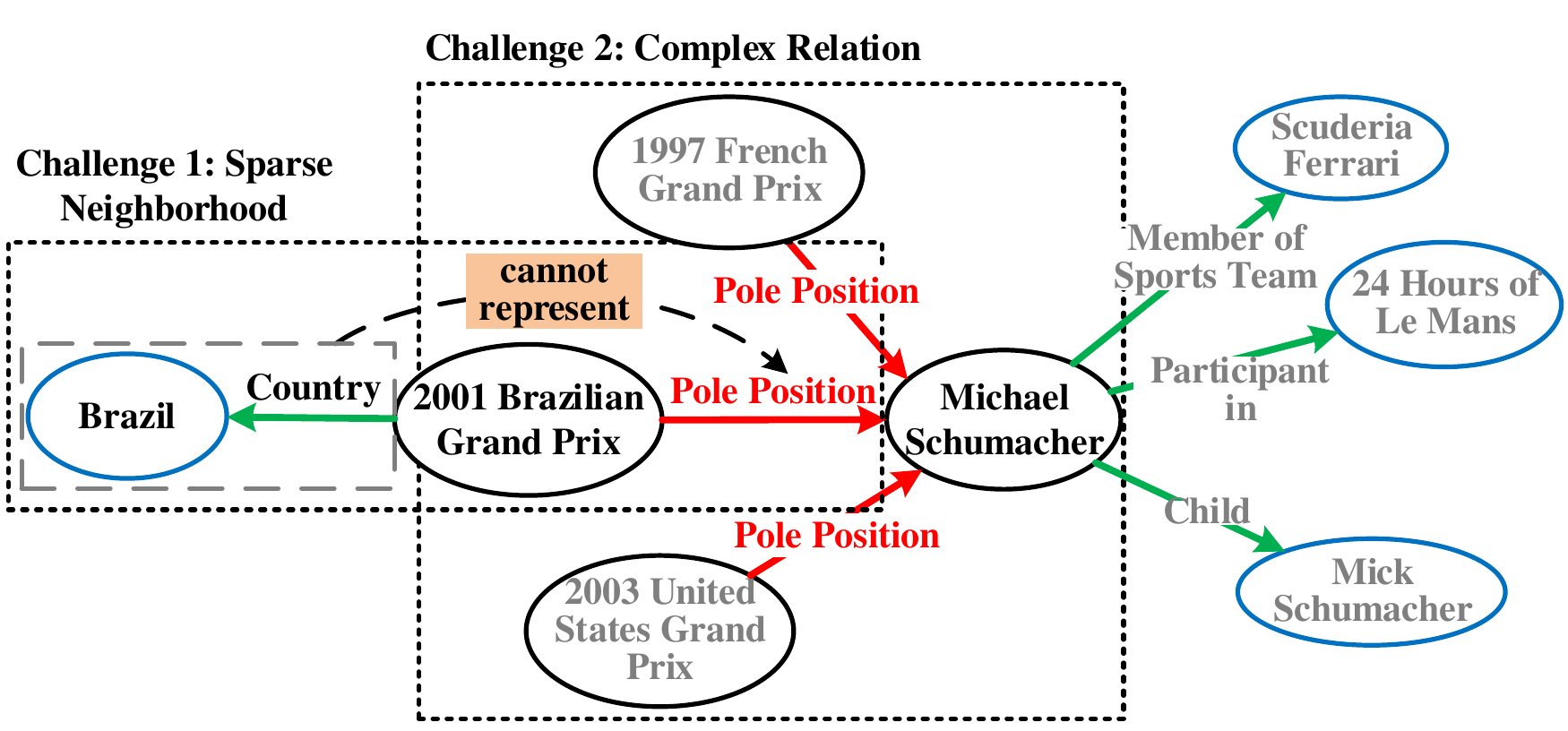}
  \caption{Two examples that exhibit the challenges of sparse neighborhood and complex relation about FKGC task on Wikidata. The edges in red denote the few-shot relations and the edges in green represent the neighbor relations. \textbf{Challenge 1}: for the few-shot relation \textcolor{red}{Pole Position}, the head entity has only a single neighbor which cannot represent the relation \textcolor{red}{Pole Position}. Thus, leveraging the sparse neighborhood would introduce noisy for representing the few-shot relation. \textbf{Challenge 2}: the few-shot relation \textcolor{red}{Pole Position} is a N-1 relation. It is hard to achieve satisfactory accuracy of predicting the head entity due to the uncertainty of the head entity specific to a N-1 relation.}
  \label{challenge}
\end{figure}

However, few-shot relations widely exist in KGs because of some natural characteristics of KGs. The first characteristic is the long-tailed distribution of the relations in a real-world KG. For instance, about 10\% of relations in Wikidata have no more than 10 triples \cite{MetaR}. Besides, for many practical applications such as social media, recommender systems, and computational finance, KGs are often updated dynamically over time. In terms of the few-shot relations in real-world KGs, their representations cannot be effectively learned with limited training instances, which causes an unsatisfactory performance on KGC.

To implement KGC in a KG with few-shot relations, FKGC models have been proposed in recent years. GMatching \cite{GMatching} and FSRL \cite{FSRL} conduct FKGC by developing a metric learning algorithm with entity embeddings and the local graph information of the entities encoded by R-GCN \cite{Michael:GNN} to compute the similarity between different entity pairs. FAAN \cite{FAAN} introduces an adaptive attentional network containing an adaptive neighbor encoder and an adaptive query-aware aggregator for FKGC by learning adaptive entity and reference representations. Meta-KGR \cite{Meta-KGR} and MetaR \cite{MetaR} both leverage model agnostic meta-learning (MAML) for fast adaption of policy network parameters or relation meta. Nevertheless, GMatching, FSRL, and Meta-KGR, focus more on the similarity or path finding rather than learning the core information for FKGC, namely the representation of few-shot relations. MetaR is not able to well handle complex relations of 1-N, N-1, and N-N, as it neglects the useful semantics in neighborhoods.

To overcome the limitations of existing methods, we propose to leverage the valuable neighbor information to represent each few-shot relation, and then use representation to infer missing facts according to the KGE models. Such a framework is much easier to optimize, and can use more contextual semantics to better deal with complex relations. However, a few challenges still remain: (1) \textbf{Sparse neighborhood}: Sparse neighborhood implies that only few relation-entity pair (neighbor) exists in the neighborhood of a few-shot relation’s head or tail entity. It will introduce the noise information when any neighbor is unable to represent the few-shot relation. In Figure \ref{challenge}, $Pole\ Position$ is a few-shot relation. The relation’s head entity only has one neighbor $(Country, Brazil)$, namely the sparse neighborhood, which cannot represent the few-shot relation $Pole\ Position$. (2) \textbf{Complex relations}: Complex relations have different categories of 1-N, N-1, and N-N.
For the tail entity $Michael\ Schumacher$ shown in Figure \ref{challenge}, multiple head entities are linked by the same relation $Pole\ Position$. Training those models requires a large amount of instances, which are unavailable to the few-shot case. Thus, these KGE models are hard to be employed directly in our setting. Although TransH \cite{Wang:TransH}, TransR \cite{Lin-a:TransR} and some other KGE methods with higher complexity could handle the complex relations, the parameters in these models are learned by a large amount of instances. Thus, these KGE models are hard to be employed immediately in the few-shot scenario.



To address the above challenges, we propose a few-shot relational learning model based on a gated and attentive neighbor aggregator, together with a MTransH for FKGC. In specific, we encode a few-shot relation neighborhood via the gated and attentive neighbor aggregator to obtain the general representation of the few-shot relation. For eliminating noise neighbor information due to the sparse neighborhood, the head and tail entities associated with the few-shot relation and their neighborhoods are combined. A gating mechanism could determine the importance of the neighborhood representation to represent a few-shot relation. We also introduce meta-learning into KGE model TransH (MTransH), benefiting for modeling the complex relations by learning the relation-specific hyper-plane parameters in the few-shot scenario. Extensive experiment results illustrate that our proposed model outperforms other state-of-the-art baselines. The source code is available at \url{https://github.com/ngl567/GANA-FewShotKGC}.

In summary, this paper makes the following contributions: 

\begin{itemize}

\item To the best of our knowledge, we are the first to propose a gated and attentive neighbor aggregator to capture the most valuable contextual semantics of a few-shot relation, combining the information from its head and tail entities as well as the entities' neighborhoods by the gating mechanism.

\item We are the first to incorporate MAML with TransH to conduct FKGC and naturally take the complex relations of 1-N, N-1, and N-N into account simultaneously.

\item The experimental results illustrate that our proposed model outperforms other KGC approaches, including the existing FKGC models. Compared with the state-of-the-art baseline MetaR, our model obtains 5-shot KGC performance improvements of 8.3\%/8.0\%/8.7\%/7.8\% on NELL-One evaluated by metrics MRR/ Hits@10/Hits@5/Hits@1. The ablation study demonstrates the effectiveness of each key module in our approach. We will release the source codes of our work.

\end{itemize}

\section{Related Work}

\subsection{Traditional Knowledge Graph Completion}

On account of the symbolic nature of the knowledge graph, logical rules can be employed for KG inference. Many rule mining tools such as AMIE+ \cite{Galarrage:AMIE} and RLvLR \cite{RLvLR} extract logical rules from the KG, which could be leveraged to infer new facts. Although the KGC approaches using logical rules have high accuracy, they cannot guarantee satisfactory generalization. In recent years, KGE \cite{KGE:Survey} techniques have been widely studied from a representation learning perspective, which could automatically explore the latent features from facts stored in a KG and learn the distributed representations for entities and relations. The well-known TransE method~\cite{Bordes:TransE} treats relations as translation operations between entity pairs to measure the compatibility of triples, which performs well on KGC. TransH~\cite{Wang:TransH} extends TransE and projects an entity embedding into different relation-specific hyper-planes, endowing an entity with various representations. With a similar idea, TransR~\cite{Lin-a:TransR} projects an entity embedding into spaces concerning various relations. ComplEx~\cite{Trouillon:ComplEx} embeds entities and relations into a complex vector space to handle the issue of inferring both symmetric and antisymmetric relations. ConvE~\cite{Dettmers:CNN} and ConvKB~\cite{ConvKB} leverage convolutional neural networks to capture the features of entities together with relations. KG2E ~\cite{he2015learning} models the uncertainties of data in a KG from the perspective of multivariate Gaussian distributions. He~\emph{et al.}~\cite{he2018knowledge} proposed a bayesian neural tensor decomposition approach to model the deep correlations or dependency between the latent factors in KG embeddings. However, these approaches all require a large amount of training instances, ignoring the practical KGs always involving long-tailed relations or dynamic update characteristics and cannot provide sufficient instances.

\subsection{Few-Shot Knowledge Graph Completion}

The existing FKGC models can be grouped into three categories: (1) Metric learning-based models: GMatching \cite{GMatching} is the ﬁrst research on few-shot (one-shot) learning for KGC. GMatching leverages a metric learning-based model that consists of a neighbor encoder to learn entity embeddings and a matching processor to measure the similarity between the reference triple and the query triple. FSRL \cite{FSRL} extends GMatching in a metric learning framework for FKGC and integrates the information from several triples in the support set while GMatching merely utilizes a single triple for one-shot KGC. (2) Meta learner-based models: MetaR \cite{MetaR} focuses on transferring relation-speciﬁc meta to represent and fast update few-shot relations. Meta-KGR \cite{Meta-KGR} proposes a meta-based multi-hop reasoning method, which employs meta-learning to learn the parameters of policy networks specific to each few-shot relation in a reinforcement learning framework. (3) Dual-process theory-based model: the latest category of FKGC is based on the cognitive system of human beings. Tang et al. \cite{CogKR} developed a Cognitive KG Reasoning model named CogKR, in which a summary module and a reasoning module work together to address the one-shot KGC problem. However, these existing FKGC approaches have not taken full advantage of the neighbors, that is, the triples related to the high-frequency relations are not effectively employed to learn the representations of few-shot relations.

\section{Problem Formulation}

\textit{\textbf{Definition of Knowledge Graph $\mathcal{G}$}}. A knowledge graph can be denoted as $\mathcal{G}=\{\mathcal{E}, \mathcal{R}, \mathcal{P} \}$. $\mathcal{E}$ and $\mathcal{R}$ are the entity set and the relation set, respectively. In specific, the relation set $\mathcal{R}$ contains high-frequency relations and few-shot relations. $\mathcal{P}=\{(h,r,t) \in \mathcal{E} \times \mathcal{R} \times \mathcal{E}\}$ denotes the set consisting of all the triple facts in the knowledge graph.

\noindent \textit{\textbf{Definition of Background Knowledge Graph $\mathcal{G}_b$}}. We formally represent a background knowledge graph as a set of triples associated with all the high-frequency relations.

\noindent \textit{\textbf{Definition of A Few-shot Relation's Neighborhood}}. In terms of a triple $(h,r,t)$ about a few-shot relation $r$, the neighborhood of $r$ is defined as $\{h,t,\mathcal{N}_h,\mathcal{N}_t\}$, where $\mathcal{N}_h$ and $\mathcal{N}_t$ are the sets of one-hop neighbors surrounding the entities $h$ and $t$ and can be generated from the background knowledge graph. A neighbor in $\mathcal{N}_h$ or $\mathcal{N}_t$ is composed of an outgoing relation and the associated tail entity as to the entity $h$ or $t$. In particular, we reverse all the incoming relations towards $h$ and $t$ in the neighborhood to be outgoing relations.


\noindent \textit{\textbf{Definition of FKGC task}}. An FKGC task can be formalized as: given a few triples in a support set $\mathcal{S}=\{(h_i,r,t_i)\}_{i=1}^{K}$ corresponding to a few-shot relation $r$, we focus on predicting the possible tail entities for a query set $\mathcal{Q}=\{(h_i,r,?)\}_{i=1}^{K}$. $K$ means the \textit{K-shot} KGC.

To reach the goal of FKGC, the training process is based on some tasks. Each task is defined as predicting new triples for a specific few-shot relation in the training set, which can be denoted as $\mathcal{T}_{train}=\{(\mathcal{S}_r, \mathcal{Q}_r)\}$. The test process $\mathcal{T}_{test}=\{(\mathcal{S}_q, \mathcal{Q}_q)\}$ is the task to infer new triples with just observing a few triples in the support set $\mathcal{S}_q$, corresponding to a relation that has not been seen during the training process.

\section{Methodology}

\begin{figure*}
  \centering
  \includegraphics[scale=0.25]{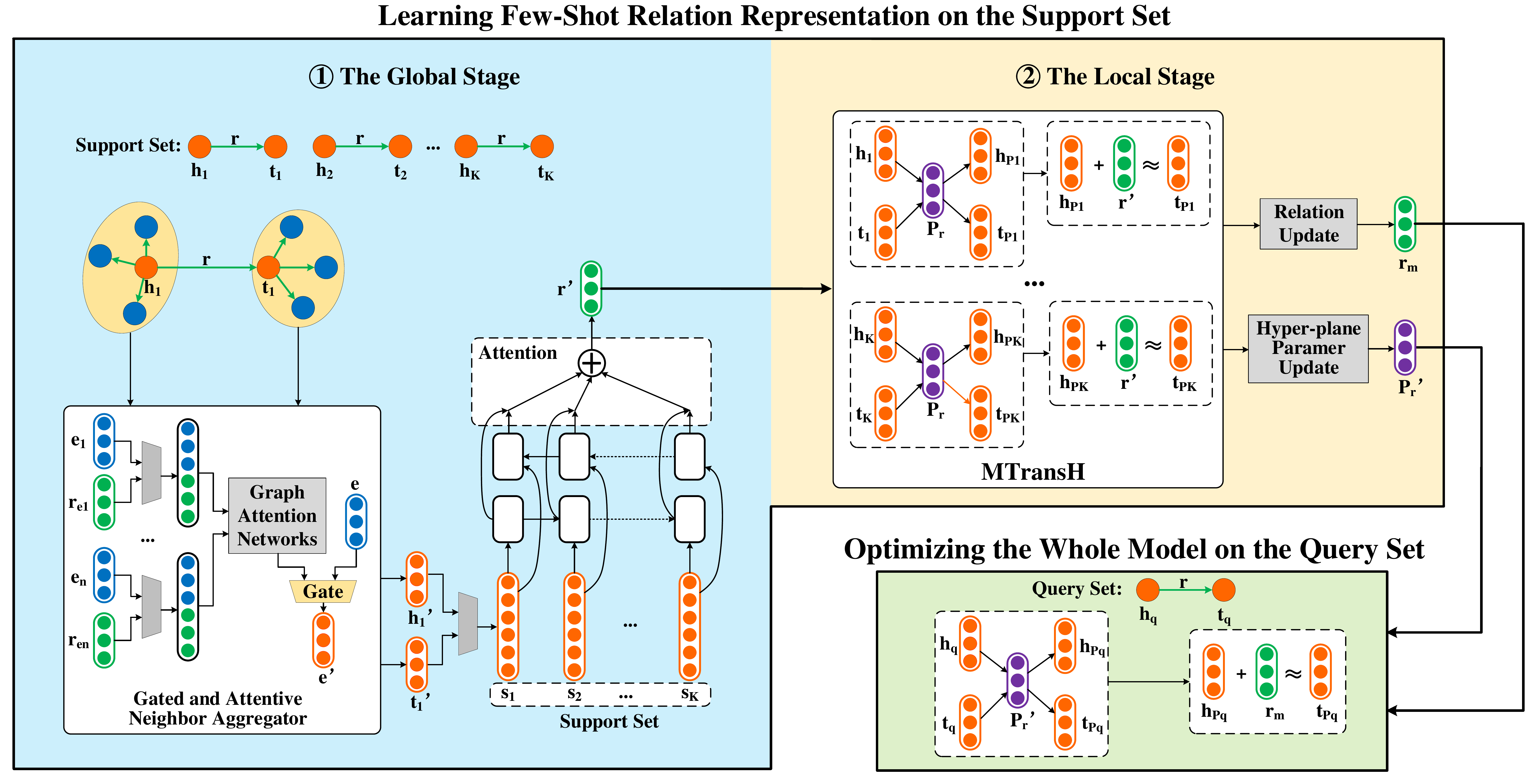}
  \caption{An overview of our model. We first learn the few-shot relation representation over the triples in the support set according to the proposed two-stage representation learning mechanism, and then optimize the whole model by the triples in the query set. We concatenate $\textbf{h}_1'$ and $\textbf{t}_1'$ as a whole vector $\textbf{s}_1$, and the vectors $\textbf{s}_2, \dots, \textbf{s}_K$ can be achieved from other triples $(h_2, r, t_2), \dots, (h_K, r, t_k)$ in the support set in the same way as obtaining $\textbf{s}_1$. $(\textbf{h}_1, \textbf{t}_1), \dots, (\textbf{h}_K, \textbf{t}_K)$ are the head and tail entity embeddings obtained by pre-train.}
  \label{figure2}
\end{figure*}

In this section, we aim to represent the few-shot relations by leveraging the neighbor information and eliminating the noisy neighbors (Challenge 1). Meanwhile, inferring complex relations in the few-shot scenario should also be addressed (Challenge 2). We will describe the detail of our proposed model. Figure \ref{figure2} shows the overview of the framework. At the global stage ($\S$ \ref{sec:GS}), on account of a triple $(h_1, r, t_1)$ about a few-shot relation $r$, the designed gated and attentive neighbor aggregator first encodes the head and tail entities with their neighbors to generate the neighborhood representation $s_1$ of the few-shot relation $r$ ($\S$ \ref{sec:GANA}). Inspired by the success of aggregating node embedding by recurrent neural network \cite{NIPS:Inductive}, all the neighborhood representations $s_1, s_2, \dots, s_K$ from the support set are integrated via an attentive Bi-LSTM to learn the general representation of the few-shot relation ($\S$ \ref{sec:LSTM}). At the global stage, we are able to learn a good initialization for the relation representations from the background knowledge graph. To further adapt the representations to the support set and derive representations which are consistent with the provided demonstrations in the support set, we further use a MAML-based approach, MTransH, to tune the representations at the local stage. More specifically, the hyper-plane parameter with respective to the relation $r$ is denoted as $\textbf{P}_r$ and the projected entity embeddings are indicated as $(\textbf{h}_{P1}, \textbf{t}_{P1}), \dots, (\textbf{h}_{PK}, \textbf{t}_{PK})$. The updated relation representation $\textbf{r}_m$ and hyper-plane parameter $\textbf{P}_r'$ can be obtained by the MTransH ($\S$ \ref{sec:local}). Finally, all the parameters can be learned by transferring the updated relation representation and hyper-plane parameter from the support set to query set.


\subsection{Global Stage: General Representation}
\label{sec:GS}

\subsubsection{Gated and Attentive Neighbor Aggregator}
\label{sec:GANA}

 Inspired by graph attention network (GAT) \cite{KBAT} that could capture different impacts of neighbors to improve entity embedding, we aim to employ a neighbor encoder based on GAT for incorporating the neighbor information of a relation to represent it. However, encoding a sparse neighborhood may even introduce noise in the process of representing a few-shot relation. Thus, a novel gated and attentive neighbor aggregator is developed to automatically capture the most valuable neighbor information for learning a few-shot relation representation and filter the neighborhood's noise information.

  In specific, given a triple $(h, r, t)$ about the few-shot relation $r$, the neighborhood of $r$ can be denoted as $\{h,t,\mathcal{N}_h,\mathcal{N}_t\}$, where $\mathcal{N}_h$ and $\mathcal{N}_t$ are the neighborhoods with respect to the entities $h$ and $t$. For the convenience of description, the head and the tail entities are both generalized as entity $e$. Then, a neighbor $(r_i, e_i)\in \mathcal{N}_e$ of the entity $e$ is encoded as followings: 
\begin{flalign}
  & \textbf{c}_i = \textbf{W}_1 \left[\textbf{r}_i;\textbf{e}_i\right] \label{eq1} \\
  & d_i = LeakyReLU(\textbf{U}_1^\mathsf{T}\textbf{c}_i) \label{eq2}
\end{flalign}
where $\textbf{r}_i$ and $\textbf{e}_i$ are the embeddings of a relation together with its linked tail entity in the neighborhood of entity $e$. $\textbf{c}_i$ is the representation of the neighbor $(r_i, e_i)$. $\textbf{W}_1$ is a linear transformation matrix and $\textbf{U}_1$ denotes a weight vector followed by a $LeakyReLU$ activation function \cite{LeakyRelu}. $d_i$ is the absolute attention value of the $i$-th neighbor.

To obtain the attention value corresponding to each neighbor, the $softmax$ function is applied over $d_i$, as shown in the followings:
\begin{flalign}
  \alpha_i &= softmax(d_i) = \frac{\exp(d_i)}{\sum_{(r_i,e_i)\in{\mathcal{N}_e}}{\exp(d_i)}}
    \label{eq3} 
\end{flalign}
where $\mathcal{N}_e$ denotes the neighborhood of the entity $e$. For eliminating the noise neighbors due to the sparse neighborhood, it is necessary to employ a gate value $g$ for automatically determining the degree of activating the neighborhood of the entity $e$ for representing the few-shot relation:
\begin{flalign}
  g = sigmoid\left(\textbf{U}_2^\mathsf{T} \sum_{(r_i,e_i)\in{\mathcal{N}_e}}{\alpha_i \textbf{c}_i} + b_g\right) \label{eqgate}
\end{flalign}
where the gate value $g$ is obtained by a linear transformation parameterized by a weight vector $\textbf{U}_2$, and a scalar bias parameter $b_g$.

Based on the gating mechanism, the representation of a half neighborhood of the few-shot relation $r$ is obtained by combining the neighbor representations of the head or tail entity $e$  with its entity embedding,
\begin{flalign}
  \textbf{e}' = \sigma\left(\sum_{(r_i,e_i)\in{\mathcal{N}_e}}{g \alpha_i \textbf{c}_i} + (1-g) \textbf{W}_2 \textbf{e} + \textbf{b} \right)  \label{eq-neighbor}
\end{flalign}
in which $\textbf{e}'$ denotes half of the neighborhood representation of the few-shot relation. $\textbf{W}_2$ denotes a weight matrix and $\textbf{b}$ is a vector parameter of bias.

Then, two neighborhood representations $\textbf{h}'$ and $\textbf{t}'$ corresponding to the head entity $h$ and the tail entity $t$ can be calculated by Eq. \ref{eq-neighbor}, and they are further concatenated together as
\begin{flalign}
  \textbf{s} = \left[\textbf{h}';\textbf{t}'\right] \label{eq5}
\end{flalign}
where $\textbf{s}$ is the neighborhood representation of the few-shot relation from the triple $(h,r,t)$ in the support set.

\subsubsection{Generating the Few-shot Relation Representation}
\label{sec:LSTM}

MetaR \cite{MetaR} represents the few-shot relation by merely averaging the relation metas from all the entity pairs in the support set, ignoring the different impacts of the triples in the support set when representing a few-shot relation. Thus, we introduce an attentive Bi-LSTM encoder for integrating multiple neighborhood representations of a few-shot relation in the support set, which can be utilized to learn the general representation of the few-shot relation. In specific, the neighborhood representations $\textbf{s}_{i}^{K}$ derived from Eq. \ref{eq5} are sequentially fed into an attentive Bi-LSTM, where $K$ is the size of the support set. The hidden states of the attentive Bi-LSTM in forward and reverse directions are computed by:
\begin{flalign}
  &\overrightarrow{\textbf{p}}_i = LSTM(W_{h1} \overrightarrow{\textbf{p}}_{i-1} + W_{s1} \overrightarrow{\textbf{s}}_{i}) \label{eq6} \\
  &\overrightarrow{\textbf{p}}_i = LSTM(W_{h2} \overleftarrow{\textbf{p}}_{i-1} + W_{s2} \overleftarrow{\textbf{s}}_{i}) \label{eq7}
\end{flalign}
in which $\overrightarrow{\textbf{p}}_{i}$ and $\overrightarrow{\textbf{p}}_{i-1}$ are hidden states of the Bi-LSTM at step $i$ and $i-1$ in the forward direction, and $\overleftarrow{\textbf{p}}_{i}$ and $\overleftarrow{\textbf{p}}_{i-1}$ are hidden states in the reverse direction. $W_{h1}$, $W_{h2}$, $W_{s1}$, $W_{s2}$ denote the parameter matrices of the Bi-LSTM. $\overrightarrow{\textbf{s}}_{i}$ is the $i$-th neighborhood representation and $\overleftarrow{\textbf{s}}_{i}$ is the $(K-i+1)$-th neighborhood representation of the input. Particularly, we select rectifier linear unit as activation function for its best performance compared to other commonly used activation functions in our experiment. The final hidden states of forward and reverse directions in $i$-th step are concatenated as a whole vector $\textbf{p}_i$, which can be further encoded as $\textbf{p}_i'$:
\begin{flalign}
  \textbf{p}_i &= \left[\overrightarrow{\textbf{p}}_i;\overleftarrow{\textbf{p}}_i\right] \label{eq8} \\
  \textbf{p}_i' &= \textbf{W}_3 \textbf{p}_i \label{eq8-1}
\end{flalign}
where $\textbf{W}_3$ is a linear transformation matrix.

Then, we calculate the weight of each final hidden state by the attention mechanism:
\begin{flalign}
  o_i &= \tanh(\textbf{U}_3^\mathsf{T} \textbf{p}_i' + b_a) \label{eq9} \\
  \beta_i &= softmax(o_i) = \frac{\exp(o_i)}{\sum_{i=1}^{K}{\exp(o_i)}} \label{eq:softmax}
\end{flalign}
where $\textbf{U}_3$ is a weight vector and $b_a$ denotes a scalar bias parameter. $\beta_i$ indicates the weight of $i$-th neighborhood representation for learning the general representation of the few-shot relation.

The final hidden states of the Bi-LSTM are summed by weight:
\begin{flalign}
  \textbf{r}' = \sum_{i=1}^{K}{\beta_i \textbf{p}_i'} \label{eq10}
\end{flalign}
where $\textbf{r}'$ denotes the general representation of the few-shot relation $r$ by integrating all the neighborhood representations from $K$ triples in the support set.

\subsection{Local Stage: MTransH}
\label{sec:local}

We aim to update the representation of few-shot relations at the local stage while taking the complex relations of 1-N, N-1, and N-N into consideration. Motivated by the typical KGE model TransH \cite{Wang:TransH} that could model the complex relations, the score function concerning a triple $(h_i, r, t_i)$ is designed as
\begin{flalign}
  &\textbf{h}_{pi}=\textbf{h}_i - \textbf{h}_i \textbf{P}_r \textbf{h}_i,\ \ \textbf{t}_{pi}=\textbf{t}_i - \textbf{t}_i \textbf{P}_r \textbf{t}_i \label{eq11} \\
    &E(h_i,r,t_i)=\Vert \textbf{h}_{pi} + \textbf{r}'-\textbf{t}_{pi} \Vert_{L1/L2} \label{eq12}
\end{flalign}
where $\textbf{h}_i$ and $\textbf{t}_i$ are the head, and tail embeddings learned in the pre-training. $\textbf{P}_r$ is the normal vector of the hyper-plane concerning the relation $r$. $\textbf{r}'$ is the general representation of the few-shot relation derived from Eq. \ref{eq10}. Based on the score calculated in Eq. \ref{eq12}, the loss function is given as
\begin{align}
    L(S_r) = &\sum_{(h_i,r,t_i)\in S_r}{\sum_{(h_i,r,t_i')\in S_r'}{\max\big[0, E(h_i, r_i, t_i) + \gamma}} - E(h_i, r_i, t_i')\big] 
    \label{eq13}
\end{align}
where $(h_i,r,t_i')$ is a triple in the set of negative samples $S_r'$ that is generated by negative sampling corresponding to the triple $(h_i,r,t_i)$.

Furthermore, we calculate the gradient of relation representation $\textbf{r}'$ by the derivative of $ L(S_r)$ with respect to $\textbf{r}'$:
\begin{equation}
  Grad(r') = \frac{\mathrm{d}L(S_r)}{\mathrm{d}r'}
  \label{eq14}
\end{equation}

Then, the relation representation can be updated following the stochastic gradient descent as follows:
\begin{equation}
  r_m = r' - l_rGrad(r')
  \label{eq15}
\end{equation}
where $l_r$ indicates the learning rate when updating the relation representation.

In particular, we leverage the model agnostic meta learning (MAML) approach to learn the hyper-plane parameter $\textbf{P}_r$ for every few-shot relation by using well-initialized hyper-plane parameters $\textbf{P}_r^*$. For a hyper-plane parameter $\textbf{P}_r$ on a task $T_r$, the hyper-plane parameter becomes $\textbf{P}_r'$ when adapting to a new task $T_{r'}$. Following MAML, the updated parameter $\textbf{P}_r'$ is calculated by a (or some) gradient descent update(s) on the support set of the current task $S_r$. For instance, the parameter $\textbf{P}_r$ can be updated by a single gradient step with the learning rate $l_p$ as following:
\begin{equation}
  \textbf{P}_r' = \textbf{P}_r^* - l_p \frac{\mathrm{d}L(S_r)}{\mathrm{d}\textbf{P}_r}
  \label{eq16-1}
\end{equation}
where $l_p$ indicates the learning rate when updating parameter $\textbf{P}_r$ on the support set. 

When we obtain the updated relation representation and hyper-plane parameter on the current task, we transfer them to instances in the query set $Q_r$. Following the same way on the support set, the score and the loss on the query set can be obtained by:
\begin{flalign}
  &\textbf{h}_{pj}=\textbf{h}_j - \textbf{h}_j \textbf{P}_{rm} \textbf{h}_j,\ \ \textbf{t}_{pj}=\textbf{t}_j - \textbf{t}_j \textbf{P}_{rm} \textbf{t}_j \label{eq17} \\
    &E(h_j,r,t_j)=\Vert \textbf{h}_{pj} + \textbf{r}_m-\textbf{t}_{pj} \Vert_{L1/L2} \label{eq18} \\
    &L(Q_r) = \sum_{(h_j,r,t_j)\in Q_r}{\sum_{(h_j,r,t_j')\in Q_r'}{max\big[0, E(h, r, t) + \gamma }} - E(h, r, t')\big] \label{eq19}
\end{flalign}
where $(h_j,r,t_j)$ is a triple in the query set $Q_r$. $L(Q_r)$ is the optimization objective for training the whole model. $(h_j,r,t_j')$ is the negative triple in the set of negative samples $Q_r'$, which is generated in the same way as $S_r'$.

Moreover, we can evaluate the updated hyper-plane parameter $\textbf{P}_r'$ on the query set of the task $T_{r'}$. Meanwhile, the well-initialized hyper-plane parameter $\textbf{P}_r^*$ can be calculated as follows:
\begin{equation}
  \textbf{P}_r^* = \textbf{P}_r - l_p \frac{\mathrm{d}L(Q_r)}{\mathrm{d}\textbf{P}_r'}
  \label{eq16-2}
\end{equation}

\begin{table*}
\renewcommand{\arraystretch}{1.5}
 \caption{Statistics of the experimental datasets.}
\centering
\begin{tabular}{c|cccccc}
\hline
Dataset		& \#Relation	& \#Entity	& \#Triples    & \#Task-Train	& \#Task-Valid	& \#Task-Test	\\
\hline
 NELL-One   & 358	        & 68,545    & 181,109    & 51	        & 5	            & 11	\\
 Wiki-One   & 822           & 4,838,244    & 5,859,240   & 133     & 16             & 34    \\
\hline
\end{tabular}
\label{table1}
\end{table*}

The detailed training procedure of our proposed model is shown as:

\begin{algorithm}  
  \caption{Training framework of our proposed model}  
  \label{alg1}
  \KwIn{
$\mathcal{T}_{train}$: Training tasks\\
\ \ \ \ \ \ \ \ \ \ $\mathcal{G}$: Background knowledge graph\\
\ \ \ \ \ \ \ \ \ \ $\Phi(N)$: Parameters of neighbor aggregator\\
\ \ \ \ \ \ \ \ \ \ $\Phi(L)$: Parameters of Bi-LSTM \\
\ \ \ \ \ \ \ \ \ \ $\Phi(P_r)$: Parameters of relation-specific hyper-plane \\
}  

  \While{not done}  
  {  
    Sample a task $\mathcal{T}_r=\{\mathcal{S}_r, \mathcal{Q}_r\}$ from $\mathcal{T}_{train}$\;
    Generate the neighborhoods of each relation in $\mathcal{S}_r$ from $\mathcal{G}$\;
    Encode the neighborhoods of each few-shot relation by Eq. \ref{eq1}-Eq. \ref{eq5}\;
    Obtain the general representation of each few-shot relation by Eq. \ref{eq6}-Eq. \ref{eq10}\;
    Calculate loss in $\mathcal{S}_r$ by Eq. \ref{eq11}-Eq. \ref{eq13}\;
    Update the relation representation by Eq. \ref{eq14} and Eq. \ref{eq15}\;
    Update $\Phi(P_r)$ by Eq. \ref{eq16-1}\;
    Calculate loss in $\mathcal{Q}_r$ by Eq. \ref{eq17}-Eq. \ref{eq19}\;
    Update $\Phi(N)$ and $\Phi(L)$ by loss in $\mathcal{Q}_r$\;
    Update the well-initialized hyper-plane parameter by Eq. \ref{eq16-2}
  }
\end{algorithm}

\section{Experiments}

In this section, the experiments for evaluating our model's performance compared with other KGC baselines, including FKGC models are conducted and their results are reported. We demonstrate the effectiveness of each key component especially the gating mechanism in our whole model by the ablation study. The visualization of the graph attention mechanism reflects different neighbors' weights when representing a few-shot relation. Besides, the results of various relation categories illustrate the superiority of MTransH.

\subsection{Datasets and Evaluation Metrics}

We evaluate our model and the baselines on two widely-used datasets NELL-One and Wiki-One for the FKGC task. Following the datasets NELL-One and Wiki-One constructed by \cite{GMatching}, the few-shot relations are selected as the relations associated with more than 50 but less than 500 triples, and other relations with their triples constitute the background knowledge graphs. The statistics of both datasets are shown in Table \ref{table1}. We utilize 51/5/11 task relations and 133/16/34 for training/validation/testing on NELL-One and Wiki-One, respectively.

To evaluate the performance of our model and the baselines for FKGC task, we utilize two traditional metrics MRR and Hits@n on both datasets. MRR is the mean reciprocal rank of the correct entities, and Hits@n is the proportion of correct entities ranked in the top n. In our experiments, n is set to 1, 5, and 10. The few-shot size is set to 1, 3, and 5 for the 1-shot, 3-shot, and 5-shot KGC tasks.

\subsection{Baselines}

We compare our model with two categories of baselines: 

(1) The traditional KGC models: including TransE \cite{Bordes:TransE}, TransH \cite{Wang:TransH}, DisMult \cite{Bordes-Weston-Bengio:semantic-matching} and ComplEx \cite{Trouillon:ComplEx}. The evaluation results of all these traditional models are achieved using the open-source codes on GitHub $\footnote{The codes for TransE, TransH, DisMult, and ComplEx are from \url{https://github.com/thunlp/OpenKE/tree/OpenKE-Tensorflow1.0}.}$. To implement FKGC for these traditional KGC models, all the triples in the background knowledge graph and the training set and the triples in the support sets about validate and test relations are employed to train the models. Then, we test these models by the triples in the query sets with respective to validate and test relations.

(2) The existing FKGC models: including GMatching \cite{GMatching}, MetaR \cite{MetaR}, CogKR \cite{CogKR} and FSRL \cite{FSRL}. The results of Hits@3 of GMatching are derived from the paper of FSRL \cite{FSRL}, and the results of the other FKGC models are obtained from their corresponding original papers. For GMatching, we use the result of pre-trained embeddings generated by ComplEx, which achieves the best results in their experiment. For MetaR, the results obtained by two kinds of settings MetaR (In-Train) and MetaR (Pre-Train) are provided.

\begin{table*}[!t]\small
\renewcommand{\arraystretch}{1.5}
\caption{Evaluation results of FKGC on NELL-One and Wiki-One. \textbf{Bold} numbers are the best results.}
\centering
\begin{tabular}{p{1.2cm}<{\centering}|p{2.2cm}<{\centering}|p{0.8cm}<{\centering}p{0.8cm}<{\centering}p{0.8cm}<{\centering}|p{0.8cm}<{\centering}p{0.8cm}<{\centering}p{0.8cm}<{\centering}|p{0.8cm}<{\centering}p{0.8cm}<{\centering}p{0.8cm}<{\centering}|p{0.8cm}<{\centering}p{0.8cm}<{\centering}p{0.8cm}<{\centering}}
\hline
    \multicolumn{2}{c|}{} & \multicolumn{3}{c|}{MRR}    & \multicolumn{3}{c|}{Hits@10}	& \multicolumn{3}{c|}{Hits@5}	& \multicolumn{3}{c}{Hits@1}\\
   
    \multicolumn{2}{c|}{\textbf{NELL-One}}    & 1-shot    & 3-shot    & 5-shot    & 1-shot    & 3-shot    & 5-shot    & 1-shot    & 3-shot    & 5-shot & 1-shot   & 3-shot    & 5-shot\\
\hline
&TransE          &  0.105        & 0.162     & 0.168         & 0.226         & 0.317     & 0.345         & 0.111         & 0.180     & 0.186      & 0.041        & 0.085  & 0.082 \\
Traditional &TransH          &  0.168        & 0.266     & 0.279         & 0.233         & 0.432      & 0.434         & 0.160         & 0.299     & 0.317      & 0.127        & 0.145  & 0.162 \\
Models &DisMult          &  0.165        & 0.201     & 0.214         & 0.285         & 0.295     & 0.319         & 0.174         & 0.223     & 0.246      & 0.106        & 0.146  & 0.140 \\
&ComplEx          &  0.179        & 0.228     & 0.239         & 0.299         & 0.345     & 0.364         & 0.212         & 0.252     & 0.253      & 0.112        & 0.165  & 0.176 \\
\hline
&GMatching       &  0.185    & 0.279     & -     & 0.313     & 0.464     & -     & 0.260     & 0.370     & -  & 0.119    & 0.198  & - \\
&FSRL            & -        & 0.318     & -         & -         & 0.507     & -         & -         & \textbf{0.433}     & -      & -        & 0.211  & - \\
FKGC &CogKR           & 0.256     & -          & -         & 0.353     & -         & -         & 0.314     & -         & -      & 0.205    & -  & - \\
Models &MetaR (In-Train)           &  0.250    & -         & 0.261     & 0.401     & -         & 0.437     & 0.336       & -         & 0.350   & 0.170    & -  & 0.168 \\
&MetaR (Pre-Train)           &  0.164    & -         & 0.209     & 0.331     & -         & 0.355     & 0.238       & -         & 0.280   & 0.093    & -  & 0.141 \\
&\textbf{Ours}	        & \textbf{0.307} & \textbf{0.322}     &\textbf{0.344} & \textbf{0.483}    & \textbf{0.510}	& \textbf{0.517}  & \textbf{0.409}  & 0.432   & \textbf{0.437}	& \textbf{0.211} & \textbf{0.225}    &\textbf{0.246}  \\
\hline
\hline
    \multicolumn{2}{c|}{\textbf{Wiki-One}}    & 1-shot    & 3-shot    & 5-shot    & 1-shot    & 3-shot    & 5-shot    & 1-shot    & 3-shot    & 5-shot & 1-shot   & 3-shot    & 5-shot\\
\hline
&TransE          &  0.036        & 0.040     & 0.052         & 0.059         & 0.072     & 0.090         & 0.024         & 0.027     & 0.057      & 0.011        & 0.013  & 0.042 \\
Traditional &TransH          &  0.068        & 0.092     & 0.095         & 1.333         & 0.170     & 0.177         & 0.060         & 0.091     & 0.092      & 0.027        & 0.045  & 0.047 \\
Models &DisMult          &  0.046        & 0.052     & 0.077         & 0.087         & 0.091     & 0.134         & 0.034         & 0.041     & 0.078      & 0.014        & 0.020  & 0.035 \\
&ComplEx          &  0.055        & 0.064     & 0.070         & 0.100         & 0.113     & 0.124         & 0.044         & 0.053     & 0.063      & 0.021        & 0.029  & 0.030 \\
\hline
&GMatching       &  0.200    & 0.171     & -     & 0.336     & 0.324     & -     & 0.272     & 0.235     & -  & 0.120    & 0.095  & - \\
&FSRL            & -        & 0.241     & -         & -         & 0.406     & -         & -         & 0.327     & -      & -        & 0.155  & - \\
FKGC &CogKR           & 0.288     & -          & -         & 0.366     & -         & -         & 0.334     & -         & -      & 0.249    & -  & - \\
Models &MetaR (In-Train)          &  0.193    & -         & 0.221     & 0.280     & -         & 0.302     & 0.233       & -         & 0.264   & 0.152    & -  & 0.178 \\
&MetaR (Pre-Train)          &  \textbf{0.314}    & -         & 0.323     & 0.404     & -         & 0.418     & \textbf{0.375}       & -         & 0.385   & \textbf{0.266}    & -  & 0.270 \\
& \textbf{Ours}	        & 0.301 & \textbf{0.331}     &\textbf{0.351} & \textbf{0.416}    & \textbf{0.425}	& \textbf{0.446}  & 0.350  & \textbf{0.389}   & \textbf{0.407} & 0.231 &\textbf{0.283}    &\textbf{0.299}  \\
\hline
\end{tabular}
\label{table2}
\end{table*}

\subsection{Implementation Details}

We use the entity and relation embeddings in the background knowledge graphs pre-trained by TransE on both datasets, released by GMatching $\footnote{\url{https://github.com/xwhan/One-shot-Relational-Learning}}$. We tune all the hyper-parameters on the validation dataset. For a fair comparison, following GMatching \cite{GMatching} and MetaR \cite{MetaR}, the embedding dimension is set to 100 for NELL-One and 50 for Wiki-One, respectively. We select the other optimal hyper-parameters by utilizing a grid search strategy based on the filter MRR on the validation dataset. The number of layers of Bi-LSTM is selected as 2, and the dimensions of Bi-LSTM's first/second layer hidden states are set to 200/100 for NELL-One and 100/50 for Wiki-One. The neighborhood's maximum size for the gated and attentive neighbor aggregator is set to 50 for both datasets. During training, we apply mini-batch gradient descent to update the parameters. The batch size is 64. The learning rates $l_r$ and $l_p$ are both selected as 0.001. The margin $\gamma$ is set to 1.0. We use the metric MRR on the validation set as the standard for the early-stop policy.

\subsection{Experimental Results and Analysis}

Three FKGC scenarios are conducted, including 1-shot, 3-shot, and 5-shot. The results on NELL-One and Wiki-One are shown in Table \ref{table2}. From the performance, we can conclude that:

\begin{itemize}

\item On both datasets, all the FKGC models consistently outperform the traditional KGC approaches, showing that few-shot relational learning effectively predicts the missing entities of few-shot relations. 

\item Our model shows consistent and significant performance improvements than all the baselines. Compared with the best baseline MetaR \cite{MetaR}, and in terms of MRR, Hits@10, Hits@5 and Hits@1, the performance gains achieved by our model are 5.1\%, 8.2\%, 7.3\% and 4.6\% on NELL-One, 8.3\%, 8.0\%, 8.7\% and 8.2\% on Wiki-One. These results illustrate that it is more effective to capture the semantics in neighbors of a few-shot relation to represent this relation.

\item Although GMatching \cite{GMatching} and FSRL \cite{FSRL} all consider the neighbor information in their model, the lack of learning few-shot relation representations in these methods limits the performance. Our model automatically selects the most relevant semantics in the neighborhood of a few-shot relation to represent it and updates the representation by MTransH. The better performance obtained by our model compared with GMatching \cite{GMatching} and FSRL \cite{FSRL} demonstrates that good representations of few-shot relations can benefit to the FKGC.

\item In terms of the results of MetaR \cite{MetaR}, only one of the two settings In-Train and Pre-Train, works well on a single dataset. Our model achieves good performance on both datasets, indicating that our approach has better generalization for different datasets.

\item Even though TransH \cite{Wang:TransH} could model the complex relations and achieve the best performance among the traditional models, it might not work for FKGC due to the limited training instances. Obviously, our model performs better than TransH \cite{Wang:TransH} and benefited from the developed gated and attentive neighbor aggregator with MTransH for few-shot relational learning.

\item  Since our model takes advantage of neighborhood to represent few-shot relations, our model has more significant performance on the datasets with richer neighborhood information such as NELL-One. On the more sparse dataset Wiki-One, our model still outperforms other baseline models and the difference between MetaR and our model shown in Table \ref{table2} is statistically significant under the paired at the 99\% significance level.

\end{itemize}

\begin{table*}[!t]
\renewcommand{\arraystretch}{1.5}
\caption{Evaluation results of 5-shot KGC on NELL-One by different categories of complex relations.}
\centering
\begin{tabular}{c|ccc|ccc|ccc|ccc}
\hline
    & \multicolumn{3}{c|}{MRR}    & \multicolumn{3}{c|}{Hits@10}	& \multicolumn{3}{c|}{Hits@5}	& \multicolumn{3}{c}{Hits@1}\\
   
    \textbf{Model}    & 1-N    & N-1    & N-N    & 1-N    & N-1    & N-N    & 1-N    & N-1    & N-N    & 1-N    & N-1    & N-N \\
\hline
TransE         & 0.095   & 0.198   & 0.088    & 0.207   & 0.397   & 0.196   & 0.112    & 0.293    & 0.163    & 0.034    & 0.186   & 0.033 \\
TransH         & 0.361   & 0.354   & 0.136    & 0.553   & 0.446   & 0.358   & 0.320    & 0.497    & 0.320   & 0.053     & 0.304   & 0.053 \\
DistMult       & 0.367   & 0.368   & 0.167    & 0.475   & 0.476   & 0.305   & 0.275    & 0.488    & 0.293   & 0.171    & 0.315   & 0.083 \\
ComplEx        & 0.392   & 0.315   & 0.131    & 0.510   & 0.347   & 0.292   & 0.414    & 0.547    & 0.222   & 0.200   & 0.320  & 0.058 \\
\hline
MetaR          & 0.169   & 0.334   & 0.207    & 0.239   & 0.480   & 0.393   & 0.345   & 0.530   & 0.320   & 0.129   & 0.312   & 0.118\\
\hline
Ours        &  \textbf{0.414}        & \textbf{0.499}     & \textbf{0.228}         & \textbf{0.598}         & \textbf{0.624}     & \textbf{0.424}         & \textbf{0.527}         & \textbf{0.572}     & \textbf{0.343}      & \textbf{0.310}        & \textbf{0.431}  & \textbf{0.120} \\
-MTransH          & 0.309        & 0.479     & 0.209         & 0.530         & 0.590     & 0.454         & 0.450         & 0.549     & 0.320      & 0.193        & 0.414  & 0.095 \\
\hline
\end{tabular}
\label{table4}
\end{table*}

\begin{table*}[!t]
\renewcommand{\arraystretch}{1.5}
\caption{Ablation Study of 5-shot KGC on NELL-One and Wiki-One.}
\centering
\begin{tabular}{l|cccc|cccc}
\hline
   & \multicolumn{4}{c|}{NELL-One}    & \multicolumn{4}{c}{Wiki-One} \\
   \textbf{Model}     & MRR    & Hits@10	& Hits@5	& Hits@1        & MRR    & Hits@10	& Hits@5	& Hits@1\\
\hline
Whole model     &  \textbf{0.344}        & \textbf{0.517}     & \textbf{0.437}     & 0.246 &  \textbf{0.351}        & \textbf{0.446}     & \textbf{0.407}     & \textbf{0.299}\\
\hline
-Gate           &  0.343        & \textbf{0.517}     & \textbf{0.437}         & \textbf{0.250}     &  0.286        & 0.436     & 0.342         & 0.230\\
-GANA           &  0.286        & 0.436     & 0.342         & 0.163     &  0.286        & 0.381     & 0.330         & 0.224\\
-MTransH        &  0.323        & 0.472     & 0.409         & 0.220     &  0.301        & 0.393     & 0.353         & 0.252 \\
\hline
\end{tabular}
\label{table3}
\end{table*}

\subsection{Results on NELL-One of Complex Relations}

We further investigate the performance of modeling the complex relations by the proposed MTransH, according to different relation categories: 1-N, N-1, and N-N. Our whole model is compared with our model that removes MTransH (-MTransH) and the traditional KGC approaches as well as the best performing baseline MetaR. The results of 5-shot KGC concerning different relation categories on NELL-One are summarized in Table \ref{table4}. We can notice that the whole model consistently and significantly outperforms the model without MTransH, illustrating the powerful ability of modeling complex relations by the developed MTransH in the few-shot scenario. Besides, we can observe that our model outperforms MetaR and other baselines including TransH which can model complex relations. MetaR even performs worse than most of the other traditional KG embedding baselines about 1-N relations due to it employs the same score function as TransE which lacks the capability of modeling complex relations. These results illustrate the effectiveness of the developed MTransH module of our model.

\subsection{Ablation Study}

Our model is a joint framework of several modules, including gating mechanism, gated and attentive neighbor aggregator, and the optimization of MTransH. To investigate the contributions of different components, we conduct 5-shot KGC and provide the following ablation studies on both NELL-One and Wiki-One from three perspectives, as shown in Table \ref{table3}:

\begin{itemize}

\item We first analyze the impact of the gating mechanism by removing it from the whole model (-Gate). From the results in Table \ref{table3}, it is evident that the gating mechanism has a remarkable influence on Wiki-One while it has little effect on NELL-One. This result is reasonable because the gating mechanism in the designed aggregator is employed to filter out the neighborhood information which cannot be used to represent the few-shot relation when the neighborhood is too sparse and even introduces noise. However, the neighborhood information of NELL-One dataset is rich enough, so the gating mechanism only plays a small role on this dataset. For Wiki-One with too sparse neighborhoods, the attention strategy in our aggregator will be useless when the neighborhood is full of noise information. At this time, the gating mechanism plays a vital role in filtering out the noise information. On the other hand, the gating mechanism makes no difference on NELL-One because the valuable information can always be found in the rich neighbors.

\item We investigate the effectiveness of our aggregator for representing few-shot relations by removing the gated and attentive neighbor aggregator (-GANA). It will cause significant performance drops, indicating the large beneﬁt of the gated and attentive neighbor aggregator to capture the valuable semantics from the neighbors of the few-shot relations.


\item We investigate the effectiveness of the MTransH. We replace the optimization objective of MTransH with TransE (-MTransH). From Table \ref{table3}, the performance drops on both datasets, indicating the superiority of incorporating MAML with TransH to model the complex relations of 1-N, N-1, and N-N for FKGC.

\end{itemize}

\begin{figure}
  \centering
  \includegraphics[scale=0.45]{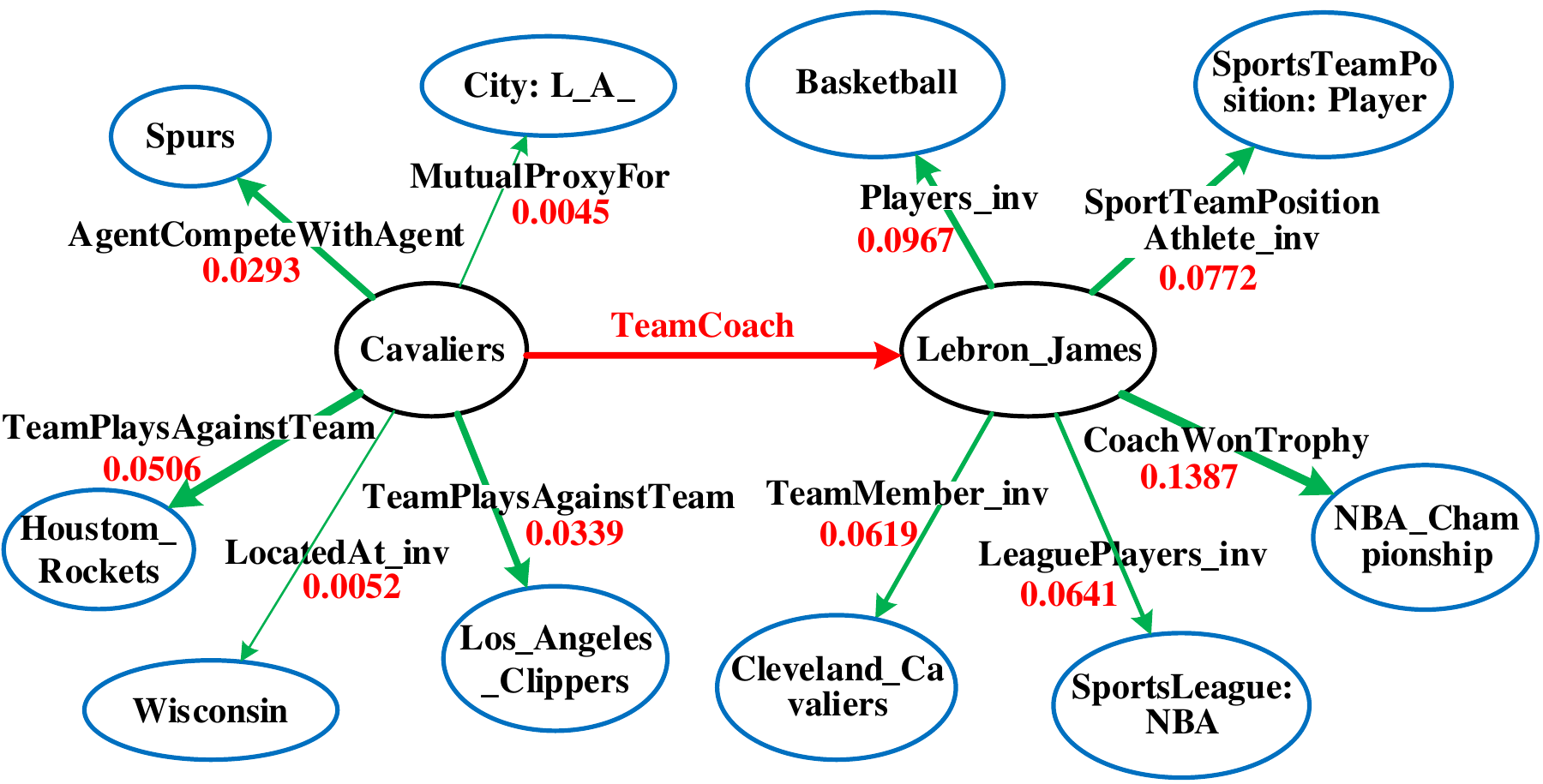}
  \caption{Visualization of the neighbors of a few-shot relation with their weights. $TeamCoach$ is a few-shot relation that connects the entity pair $(Cavaliers, Lebron\_James)$ and the others are the neighbors. We ensure the relations in the neighborhoods are all outgoing relations towards the entity pair of the few-shot relation, and reverse the incoming relations with the suffix "$inv$". We select three neighbors with the largest weights and two neighbors with the smallest weights from each neighborhood.}
  \label{figure3}
\end{figure}

\subsection{Visualization}

The key idea of our approach is representing the few-shot relations by their neighbors. Thus, we need to know how different information in the neighborhood differs in representing a few-shot relation. Figure \ref{figure3} visualizes the neighbors of a few-shot relation together with their weights calculated by the designed gated and attentive neighbor aggregator to explain the different levels of significance of the neighbor information in representing a few-shot relation more clearly. We can observe that the neighbor (\textbf{TeamPlaysAgainstTeam}, \textbf{Houstom\_Rockets}) is assigned with the largest weight for representing the concept \textbf{Team} and (\textbf{CoachWonTrophy}, \textbf{NBA\_Championship}) has the largest weight for representing the concept \textbf{Coach}, which have the most relevant semantics with the few-shot relation \textbf{TeamCoach}. Meanwhile, the neighbors (\textbf{MutualProxyFor}, \textbf{City:L\_A\_}) and (\textbf{TeamMember\_inv}, \textbf{Clevelan\_Cavaliers}) are given the smallest weights in the neighbors of the head entity and tail entity, respectively, which shows these two neighbors merely play a minor role in representing the few-shot relation. The visualization illustrates the effectiveness of the gated and attentive neighbor aggregator for representing few-shot relations.

\section{Conclusion and Future Work}

This paper presents a relational learning approach for FKGC task. Even though there are very sparse neighborhoods in a KG, the most valuable neighbors can be captured to learn the general representations of a few-shot relation by a novel gated and attentive neighbor aggregator. The few-shot relation representations can be updated via the developed MTransH, benefiting for modeling and inferring the complex 1-N, N-1, and N-N relations. The experimental results demonstrate that our model outperforms other state-of-the-art baselines on the two few-shot datasets. The ablation study and the results of complex relations by various categories illustrate the significant effects of the gated and attentive neighbor aggregator and the MTransH. In future work, we will consider more external knowledge, such as the text description of relations and entities for enhancing the representation of relations and entities. Besides, both few-shot entities and relations will be considered in the FKGC task.

\bibliographystyle{ACM-Reference-Format}
\balance
\bibliography{sample-base}

\end{document}